\documentclass[runningheads]{llncs}
\usepackage[T1]{fontenc}
\usepackage{graphicx} 
\usepackage{float}
\usepackage{algorithm}  
\usepackage{algpseudocode}
\usepackage{color}
\usepackage{setspace}
\usepackage{multirow}  
\usepackage{booktabs}
\usepackage{lineno}
\usepackage{amssymb}
\usepackage{bbding}
\begin{document}
\title{Self-supervised transformer-based pre-training method with General Plant Infection dataset}
\author{Zhengle	Wang\inst{1,4} \and
Ruifeng	Wang\inst{2} \and
Minjuan	Wang\inst{1, 4(}\Envelope\inst{)}\and 
Tianyun Lai\inst{3} \and 
Man Zhang\inst{1,4}}
\authorrunning{F. Author et al.}
%
\institute{Key Laboratory of Smart Agriculture Systems, Ministry of Education, China Agricultural University, Beijing 100083, China \and
College of Engineering, China Agricultural University, Beijing 100083, China \and
College of Economics and Management, China Agricultural University, Beijing 100083, China \and
College of Information and Electrical Engineering, China Agricultural University, Beijing 100083, China
}
\maketitle              
\begin{abstract}
Pest and disease classification is a challenging issue in agriculture. The performance of deep learning models is intricately linked to training data diversity and quantity, posing issues for plant pest and disease datasets that remain underdeveloped. This study addresses these challenges by constructing a comprehensive dataset and proposing an advanced network architecture that combines Contrastive Learning and Masked Image Modeling (MIM). The dataset comprises diverse plant species and pest categories, making it one of the largest and most varied in the field. The proposed network architecture demonstrates effectiveness in addressing plant pest and disease recognition tasks, achieving notable detection accuracy. This approach offers a viable solution for rapid, efficient, and cost-effective plant pest and disease detection, thereby reducing agricultural production costs. Our code and dataset will be publicly available to advance research in plant pest and disease recognition the GitHub repository at \url{https://github.com/WASSER2545/GPID-22}

\keywords{Self-supervised model  \and Large Pest and Disease dataset \and Pest and disease classification \and Mask Image modeling \and Contrastive learning}
\end{abstract}

\section{Introduction}
\label{intro}
It is widely acknowledged that the main causes of crop damage include both biotic and abiotic stresses, among which plant pests and diseases are the primary factors adversely affecting crop yield and quality
 \cite{1.1}. 
Reports in \cite{1.4} and \cite{1.3} indicates that over $80\%$ of agricultural products worldwide are produced by farmers, while yield losses due to plant pests and diseases exceed $50\%$. Specifically, wheat, rice, maize, potatoes, and soybean yields have decreased by $21.5\%$, $30.0\%$, $22.5\%$, $17.2\%$, and $21.4\%$, respectively.
Hence, the classification of pests and diseases is crucial in agricultural production, playing a vital role in ensuring food security
and stabilizing the agricultural economy
. Unfortunately, the wide variety of plant pests and diseases, along with the minor differences between different plant diseases and pests, mean that traditional identification methods, which heavily rely on manual inspection by agricultural experts \cite{1.1}, are costly, labor-intensive, and prone to human error, leading to low classification efficiency and unreliable results \cite{1.6}. This significantly hinders the development of agricultural production, underlining the importance of researching new, cost-effective, and efficient methods for plant pest and disease identification.

The development of deep learning and computer vision technologies has significantly advanced precision and smart agriculture
, notably enabling automated recognition of plant pests and diseases.

Traditional detection methods mainly relied on machine learning frameworks with Manually Designed Features and Classifiers (or Rules) \cite{1.8}. These methods included 
\cite{1.9,1.12,1.11}(1) Image Segmentation Methods: threshold segmentation; Roberts, Prewitt, Sobel, Laplace and Kirsh edge detection; region segmentation; (2) Feature Extraction Method: SIFT, HOG, LBP, shape, color and texture feature extraction method; (3) Classification Method: SVM, BP, Bayesian.

Compared to other image recognition technologies, deep learning-based image recognition does not require the extraction of specific features. It finds the appropriate features through iterative learning, capturing both global and contextual characteristics of images. This method offers strong robustness and high recognition accuracy\cite{1.8}, playing a crucial role in the core of smart agriculture. Over the past few decades, various deep learning architectures have been proposed for the classification of plant pests and diseases, leading to the development of several diagnostic systems suitable for practical agricultural production\cite{1.6}.

Guowei Dai et al. \cite{1.14} developed a 
deep learning model for plant disease recognition using the DFN-PSAN framework. Their approach redesigned YOLOv5's classification layer with PSA attention to highlight important areas of plant diseases at the pixel level.
Trained on three datasets, the model demonstrated high performance with an average accuracy and F1-score exceeding $95.27\%$.
Shang Wang et al. \cite{1.17} developed the ODP-Transformer model for plant pest detection. This model consists of a pest part detector based on the Faster R-CNN framework, a Part Sequence Encoder, a Description Decoder, and a Classification Decoder. Trained on the APTV-99 dataset, their model outperformed 8 common CNN models in plant pest image classification tasks by $12.91\%$.

Despite the widespread application of deep learning technologies in various computer vision tasks, obstacles 
still exist in practical applications
:

\begin{enumerate}
\item The performance of deep learning models is directly tied to the diversity and volume of their training data. 
However, for many plant diseases and pests, relevant data are still scarce. Insufficient data volume can lead to over-fitting, while limited data diversity may affect the models' robustness. Moreover, most of the code and datasets from existing research are not publicly available, resulting in models that may perform poorly, have unknown robustness, and are difficult to adapt to other application scenarios.


\item Previous research on plant pest and disease classification has largely utilized Convolutional Neural Networks (CNNs). However, CNNs require a substantial amount of training data \cite{1.12,1.11} and retraining them can be both time-consuming and costly.

\item Subtle differences between various pests and overlapping features among different diseases can lead to reduced accuracy in recognition models.
\end{enumerate}

Addressing the outlined deficiencies, this paper aims to create a plant pest and disease dataset that balances volume and diversity. It proposes an improved network that integrates Contrastive Learning and Masked Image Modeling (MIM) to enhance the recognition, classification efficiency, and reliability of plant pests and diseases, thereby reducing agricultural production costs. Our contributions are summarized as follows:

\begin{enumerate}
    \item We 
    created a dataset that includes common plant pests and diseases, incorporating most of the open-source datasets 
    online for plant disease classification
    . Our dataset, which is the largest in terms of sample volume and the variety of plant pests and diseases included to our knowledge, comprises images of 183 types of pests and diseases across 22 plant species, totaling 205,371 images, from 199 different classes. And we plan to make our code and 
    dataset open source to the scientific community, aiming to foster further research in plant pest and disease recognition.
    \item We developed an architecture based on the Vision Transformer (VIT) network with 
    MIM, incorporating contrastive learning during pre-training. Our experimental results indicate that this method performs effectively in recognizing plant pests and diseases.
    \item Based on the GPID-22 method proposed in this paper, our approach achieved state-of-the-art results for 
    199 different classes. This method offers a viable solution for fast, efficient, and low-cost detection of plant pests and diseases.
\end{enumerate}

\section{Materials and Methods}
\subsection{Datasets}
\subsubsection{Data Collection and Data Annotation}
\begin{figure}[H]
    \centering
    \includegraphics[height=4.8cm]{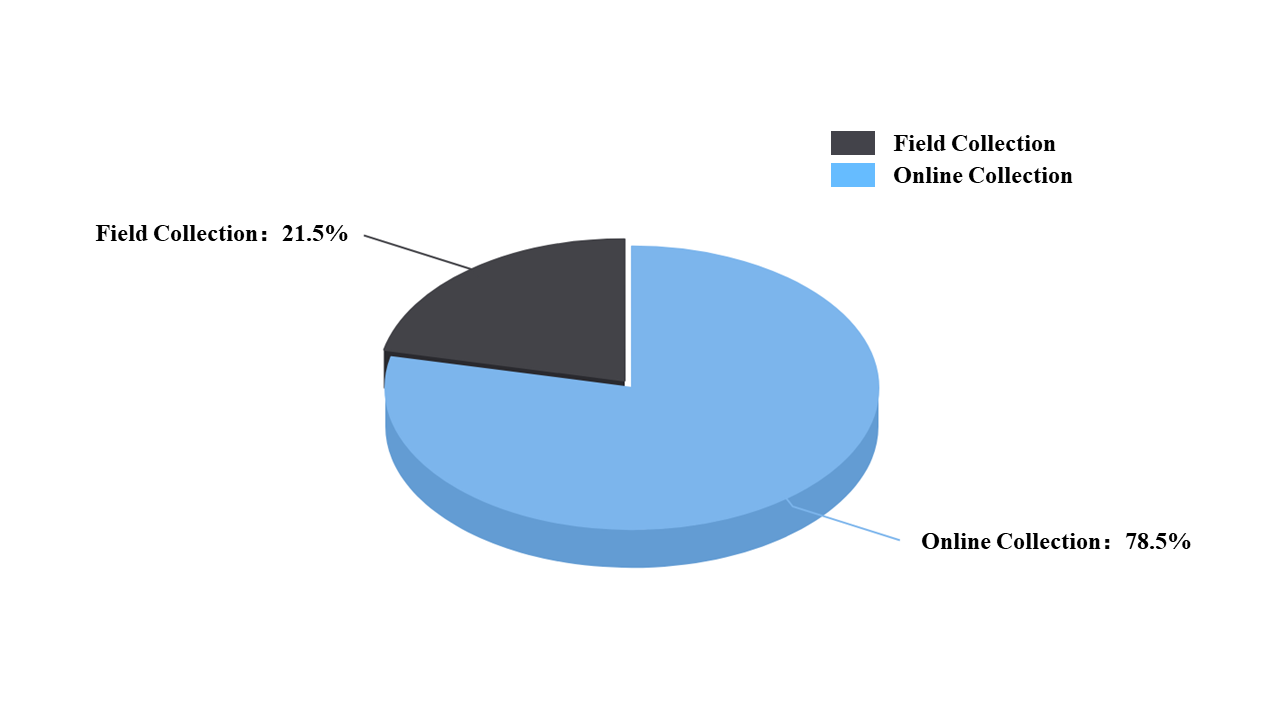}
    \caption{The sample source proportions in the GPID-22 dataset}
    \label{fig:xa}
\end{figure}

\begin{figure}[H]
    \centering
        \includegraphics[height=3.8cm]{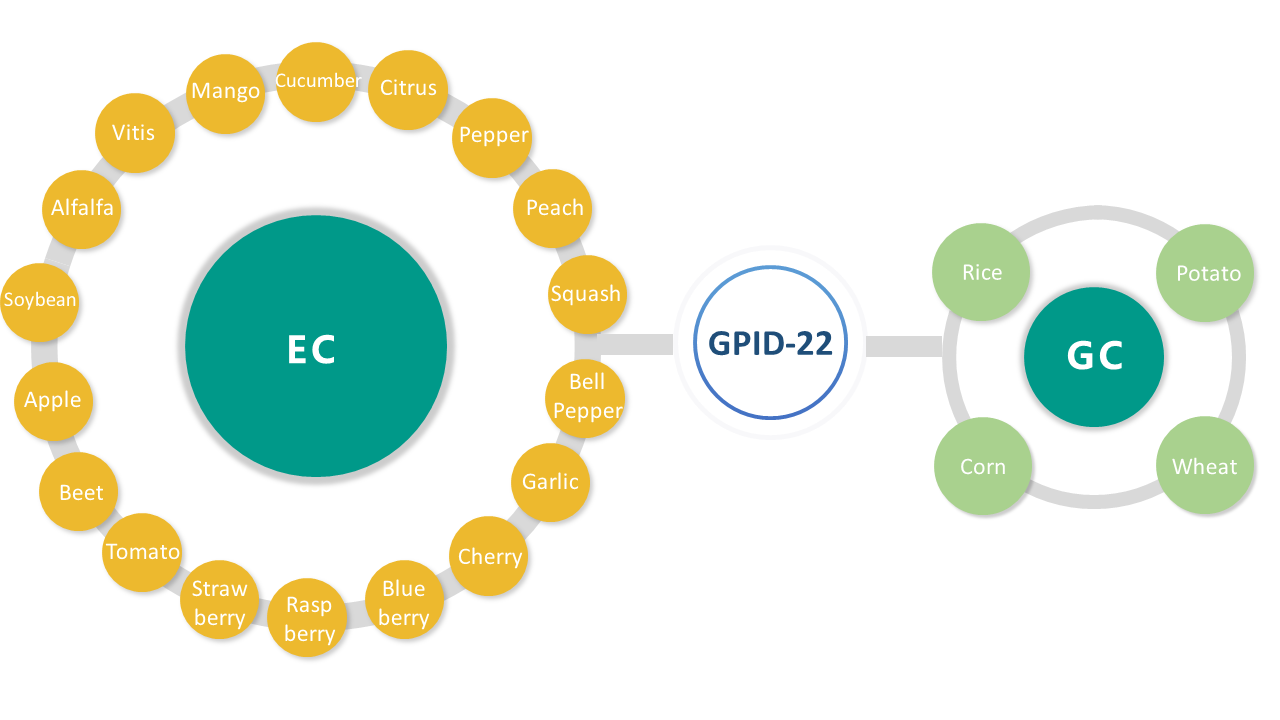}
    \caption{Taxonomy of the GPID-22 dataset. "EC" and "GC" respectively stand for Economic Crops and Grain Crops. In this context, we have only displayed the base-classes within our classification system. The complete list of sub-classes is available in our openly accessible dataset.}
    \label{fig:xb}
\end{figure}

We collected and annotated our dataset through the following five stages: (1) Collection of 
online images, (2) Establishment of a classification system, (3) Collection of field images, (4) Preliminary data filtering, (5) Expert data annotation.
In the collected data, $78.5\%$ of the images are from other datasets, and $21.5\%$ 
are collected from the field, as shown in Figure \ref{fig:xa}. For the online collection, we first conducted an extensive search for open-source datasets in well-known platforms and research databases. The datasets cited include IP102, PlantVillage, and New Plant Diseases Dataset
\cite{NPDD,plantvillage,ip102}. Additionally, we considered several factors such as the source and size of the datasets, their credibility and reputation, the number of plant diseases or pest categories included, and the consistency and accuracy of the annotations and metadata. Notably, we only referenced the training sets of these open-source datasets. To ensure the scientific rigor of our study, we cleaned the pre-training sets of the collected open-source datasets by removing low-quality images and eliminating duplicate images from different datasets.
\vspace{-0.5cm}
\vspace{-0.2cm}
Subsequently, we established a hierarchical classification system for our dataset. We conducted surveys on several important crops at the an Experimental Station, including tomatoes, potatoes, cucumbers, grapes, rice, and wheat. We also invited several agricultural experts to discuss the common pests and diseases of the above crops. Ultimately, 66 types were identified, and a hierarchical structure was constructed with the collected open-source datasets, as shown in Figure \ref{fig:xb}. Then, we collected field images of common pests and diseases of the aforementioned crops. For pests, based on their primary crops affected, we categorized each pest into an upper-level class, referred to as a base-class in the following text. That is to say, each pest is a sub-class of a specific base-class, which is referred to as a sub-class in the following text. For example, Spodoptera litura, which primarily affects tomatoes, with tomatoes belonging to Economic Crops (EC). Therefore, Spodoptera litura belongs to the base-class of tomatoes and EC. The detailed structure of our dataset will be introduced in Section 2.1.2.
\vspace{-0.5cm}

\begin{figure}[h]
    \centering
    \includegraphics[height=7cm]{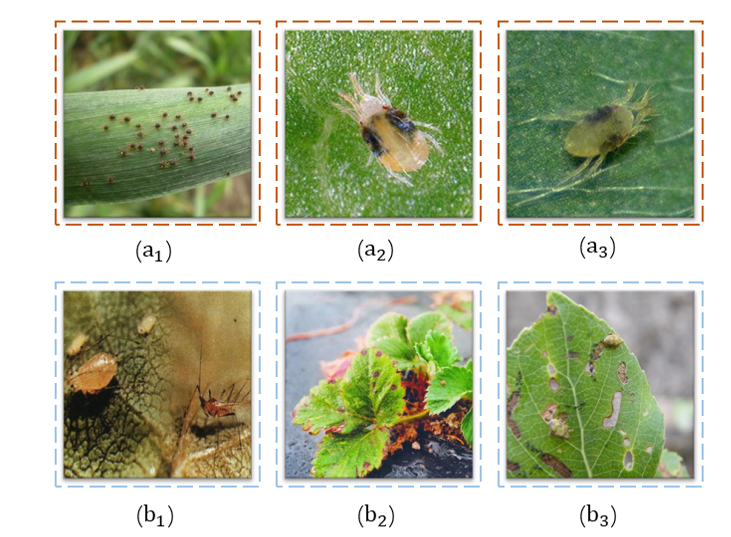}
    \caption{Different forms of insect pest images and images to be discarded}
    \label{fig:xc}
\end{figure}

We collected over 160,000 candidate images for our dataset from fieldwork and organized a team of five researchers to manually filter these candidate images. During the filtering process, researchers, who had received training on basic knowledge of plant pests and diseases, the taxonomic system of our dataset, and the different forms of plant pests and diseases (recognizing that pests at different stages of their life cycle can still cause varying degrees of damage to crops), eliminated images that contained more than one type of pest, more than one type of disease, or images that had both pests and diseases, as shown in Figure \ref{fig:xc}. Subsequently, we converted the field-collected images to JPEG format and removed any duplicates and damaged images. Ultimately, we obtained approximately 70,000 images that were weakly labeled using search keywords. Each base-class label was derived according to the classification system of our dataset.

 \vspace{-0.8cm}
 
\begin{figure}[h]
    \centering
    \includegraphics[height=9cm]{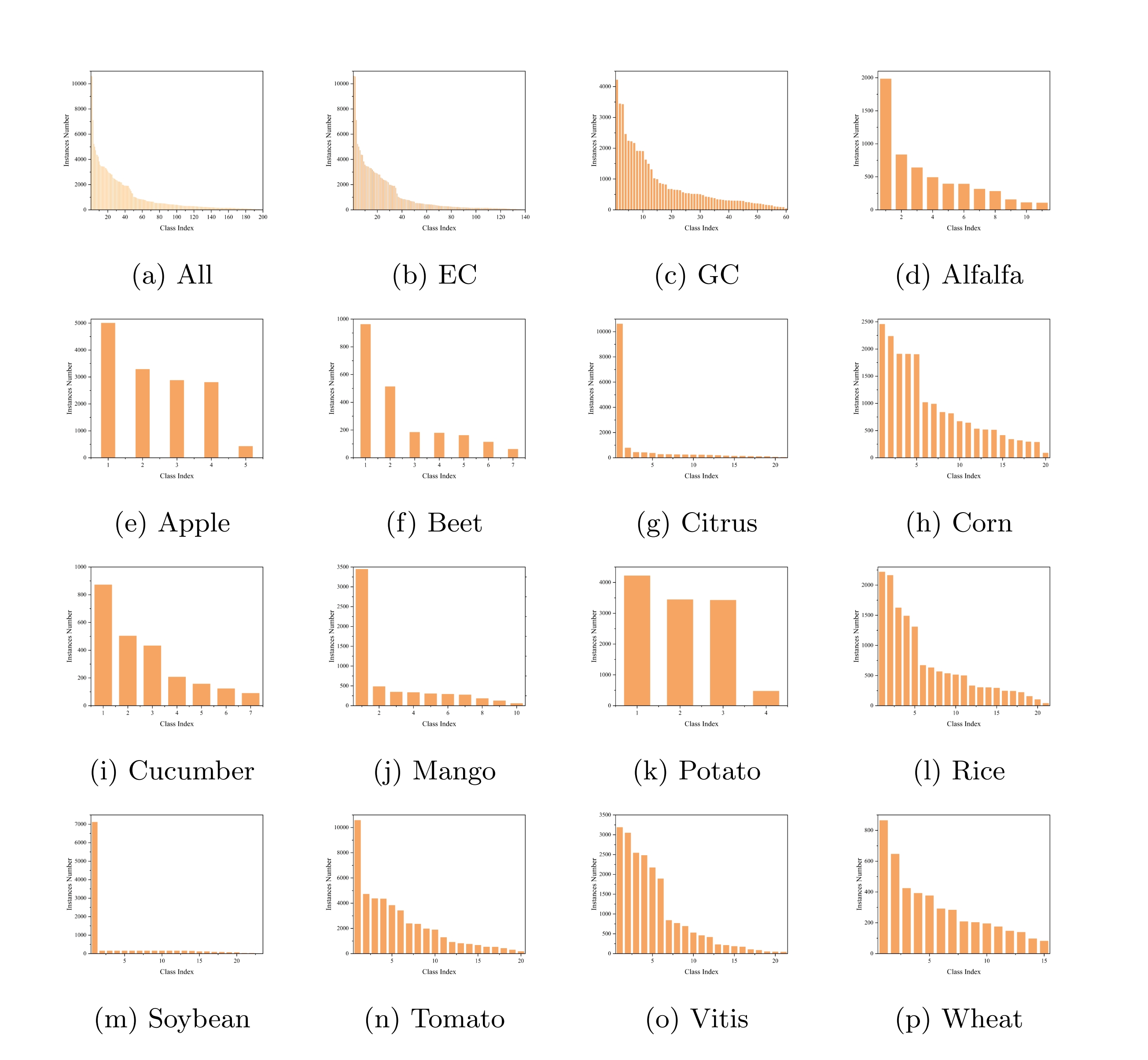}
    \caption{Sample number distribution of the GPID-22 dataset in different levels. In the diagram, only the overall situation is listed, including "EC", "GC", and 13 types of crops that are affected by pests and diseases.}
    \label{fig:xd}
\end{figure}

The data annotation by agricultural experts is the most crucial process in our dataset. In the classification system of our dataset, there are 22 crops affected by diseases and pests, with field-collected images coming from 6 types of crops. For each type of crop, we invited agricultural experts who primarily research the corresponding crops. Therefore, we invited
6 agricultural experts to annotate the images after the initial screening. Subsequently, we converted the collected open-source images to JPEG format, merged the collected open-source images with the field-collected images, and eliminated any duplicate and damaged images. Finally, we obtained a total of 44,155 images that had been annotated, with each base-class and sub-class being determined according to the classification system of our dataset. The method of combining field photography and online collection significantly increased the quantity and diversity of our dataset's data. Figure \ref{fig:xd} shows the distribution of sample numbers across different levels of our dataset.

\subsubsection{Dataset Division and Structure}
Our dataset (GPID-22) comprises 205,371 images, encompassing 199 classes across 22 types of plants, including diseases, pests, and healthy specimens, with the smallest category containing only 5 samples. To achieve more reliable test results on our dataset, each category in the test set should have a sufficient number of samples. Therefore, GPID-22 was divided approximately in an 8:2 ratio. Additionally, both the pre-training and test sets were divided according to sub-class categories. Specifically, our dataset allocated 164,700 images for pre-training and 40,671 images for testing. Detailed division tables for different levels are available in Table \ref{tab:xa}. The list of images corresponding to each set is accessible within 
GPID-22.

\begin{table}[h]
  \centering
  \caption{Pre-training/testing (denoted as Pre-train/Test) set split of the GPID-22 dataset on different class levels. The "Class" indicates the sub-class number of the corresponding base-class. "EC" and "GC" denote the economic crops and grain crops.}
  \scalebox{0.8}{
\setlength{\tabcolsep}{6.5mm}{
  \begin{tabular}{c|cccc}
    \toprule
    \multicolumn{2}{c}{Base-class} & Class & Pre-Train & Test \\
    \hline
    \multirow{18}[2]{*}{EC} & \multicolumn{1}{l}{Citrus} & 21    & 12274 & 3068 \\
          & \multicolumn{1}{l}{Mango} & 10    & 4672  & 1168 \\
          & \multicolumn{1}{l}{Vitis} & 21    & 16107 & 4027 \\
          & \multicolumn{1}{l}{Alfalfa} & 11    & 4562  & 1141 \\
          & \multicolumn{1}{l}{Soybean} & 23    & 7690  & 1923 \\
          & \multicolumn{1}{l}{Apple} & 5     & 11512 & 2879 \\
          & \multicolumn{1}{l}{Beet} & 7     & 1741  & 435 \\
          & \multicolumn{1}{l}{Tomato} & 20    & 36961 & 9240 \\
          & \multicolumn{1}{l}{Bell Pepper} & 2     & 4303  & 1076 \\
          & \multicolumn{1}{l}{Pepper} & 2     & 1351  & 338 \\
          & \multicolumn{1}{l}{Peach} & 2     & 6506  & 1627 \\
          & \multicolumn{1}{l}{Strawberry} & 2     & 4211  & 1053 \\
          & \multicolumn{1}{l}{Squash} & 1     & 2857  & 714 \\
          & \multicolumn{1}{l}{Raspberry} & 1     & 2225  & 556 \\
          & \multicolumn{1}{l}{Cherry} & 2     & 3178  & 794 \\
          & \multicolumn{1}{l}{Garlic} & 1     & 223   & 56 \\
          & \multicolumn{1}{l}{Blueberry} & 1     & 2654  & 566 \\
          & \multicolumn{1}{l}{Cucumber} & 7     & 1907  & 477 \\
    \hline
    \multirow{4}[2]{*}{GC} & \multicolumn{1}{l}{Corn} & 20    & 14942 & 3735 \\
          & \multicolumn{1}{l}{Rice} & 21    & 11560 & 2890 \\
          & \multicolumn{1}{l}{Wheat} & 15    & 3618  & 904 \\
          & \multicolumn{1}{l}{Potato} & 4     & 9242  & 2311 \\
    \hline
    \multicolumn{2}{c}{Our Dataset} & 199   & 164700 & 40671 \\
    \bottomrule
  \end{tabular}}}
  \label{tab:xa}%
\end{table}%

In Table \ref{tab:xb}, we compared 
GPID-22 with several existing datasets relevant to the task of plant disease or pest identification. Compared to the IP102 training set \cite{ip102}, which contains 45,095 images, GPID-22's Pre-training set includes approximately 3.65 times more samples with 164,700 images. Relative to the PlantVillage training set \cite{plantvillage}, which has 43,442 images, our dataset's pre-training set contains about 3.79 times the number of samples. When compared to the New Plant Diseases Dataset training set \cite{NPDD}, which includes 70,295 images, GPID-22's pre-training set has about 2.34 times the number of samples. In terms of plant species diversity, the IP102, PlantVillage, and New Plant Diseases Dataset only include 8, 14, and 14 types of affected plants, respectively, whereas our dataset encompasses 22 types of plants affected by diseases or pests. Regarding the diversity of diseases and pests, IP102, PlantVillage, and New Plant Diseases Dataset include only 102, 27 (38 classes in total), and 27 (out of 38 different classes) types of pests or diseases, respectively, while GPID-22 boasts 183 types of plant pests or diseases (out of 199 different classes). In terms of the average number of samples per class, our dataset has at least 384 more samples than IP102. Beyond these statistical differences, currently, only a portion of datasets are open source, with only PlantVillage, IP102, and New Plant Diseases Dataset being of significant size. Limited by a scarcity of samples, insufficient sample diversity, and the lack of open access, most datasets related to plant diseases and pests 
\cite{3.3,3.4,3.1} are challenging to apply in practical scenarios.

\begin{table}[h]
  \centering
    \caption{Comparison with existing datasets related to plant diseases and pests. The "Class" denotes the class number. The "AVA" indicates if the dataset is available. The "Y" and "N" denote "Yes" and "No", respectively. "Samp" indicates the number of samples in training set,The ‘Avg’ denotes average numbers of samples per class.}
    \setlength{\tabcolsep}{3mm}{
    \begin{tabular}{cccccc}
    \toprule
    Dataset & Year  & Class & Ava   & Samp  & Ave \\
    \midrule
    IP102 \cite{ip102} & 2019  & 102   & Y     & 45095 & 442 \\
    PFIP \cite{3.1} & 2014  & 20    & N     & 160   & 8 \\
    Pest ID \cite{3.3} & 2016  & 12    & N     & 3595  & 300 \\
    Tea Insect Pests Database \cite{3.4} & 2012  & 8     & N     & 424   & 53 \\
    GPID22 & 2024  & 199   & Y     & 164297 & 826 \\
    \bottomrule
    \end{tabular}}

  \label{tab:xb}
\end{table}

\subsection{Method}
\begin{figure}[H]
\centering
\includegraphics[width=0.9\textwidth]{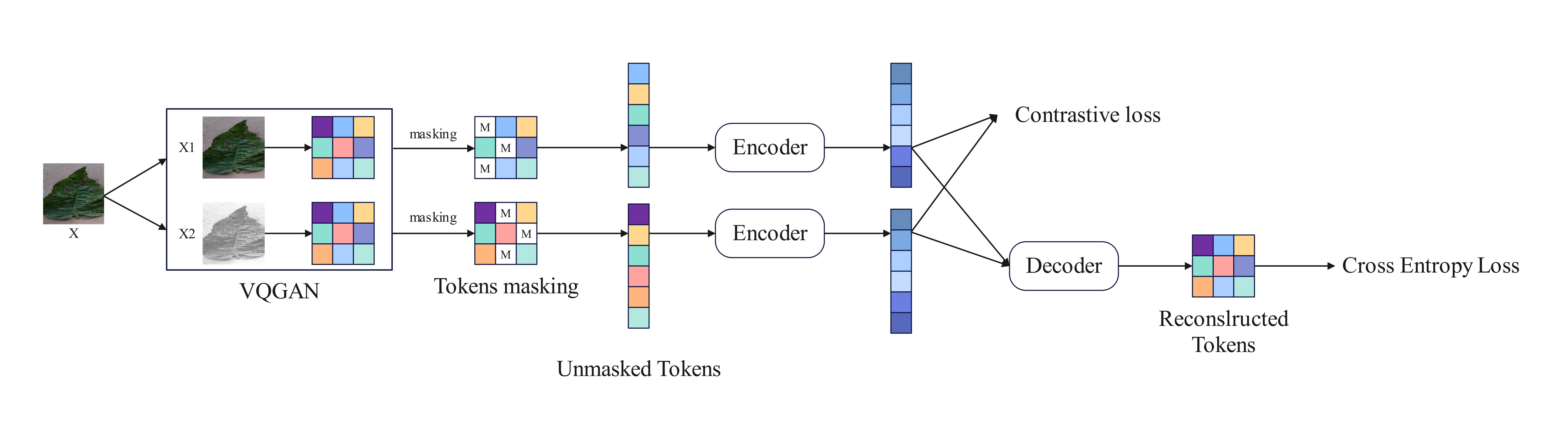}
\caption{CRE Framework}
\label{fig:f1}
\end{figure}

To enhance the generality of methods in agricultural pest and disease detection tasks and enable the model to learn more generalized image features, we propose a pre-trained model based on Masked Image Modeling (MIM). The framework of the model is presented in Figure \ref{fig:f1}. The first layer of the entire model is a pre-trained Vector Quantized Generative Adversarial Network (VQGAN) model, through which input images are processed to yield semantic tokens. These tokens are then subjected to a masking operation. Following the setup of MAGE \cite{mage}, the masked tokens are fed into an encoder-decoder transformer architecture. Finally, we added a contrastive-learning branch to the model, enabling it to learn information within the feature space, thereby enhancing its capability to handle downstream tasks more effectively.

\subsubsection{Self-supervised image reconstraction module}
In the proposed CRE architecture, we primarily utilize MAGE as the backbone for our model's reconstruction branch. To more effectively extract image features, we adopt a design similar to MAGE during the masking phase, where the image is input into a VQGAN tokenizer \cite{esser2021taming} to obtain a sequence of semantic tokens $Y=\left[y_{i}\right]_{i=1}^{L}$, with $L$ being the sequence length, and $Mask=\left[m_{i}\right]_{i=1}^{L}$ representing the boolean vector that identifies the masked tokens. We then apply the same masking ratio $r = 0.55$ as in MAGE and randomly replace $L*r$ length tokens with a mask vector $M$.

Upon obtaining the sequence of unmasked tokens, we feed it into a 
VIT \cite{ViT} encoder-decoder structure. The encoder encodes the unmasked token sequence into a latent feature space. Subsequently, we utilize learnable masking tokens to populate the sequence obtained from the encoder and input it into the ViT decoder. Finally, we input the filled token sequence into the decoder for reconstruction to obtain the reconstruction result $p(y_i)$. The reconstruction loss $\mathcal{L}_{reconstruction}$ is calculated as a cross-entropy loss between the input one-hot tokens and $p(y_{i})$.

\begin{equation}
    \label{eq:eq1}
\mathcal{L}_{reconstruction}=-\mathbb{E}_{Y\in\mathcal{D}}(\sum_{\forall i,m_i=1}\log p(y_i))
\end{equation}

\subsubsection{The contrastive learning module}
Referring to the observation in \cite{CMAE}, the inclusion of a contrastive loss in the MIM method leads to enhanced performance in representation learning. So we similarly incorporate a contrastive-learning branch to enhance the linear separability of the acquired feature space. Following 
SimCLR's approach\cite{simclr}, we generate a pair of positive samples by using random augmentations $\tau_{1}$ and $\tau_{2}$ for each input image, then we apply a two-layer MLP on the feature derived from globally averaging the encoder's output. Subsequently, we introduce an InfoNCE loss \cite{Oord_Li_Vinyals_2018} applied to the output of the MLP head \ref{eq:eq2}, where $z$ denotes the normalized features obtained after the two-layer MLP, $B$ denotes the batch size, and $t$ denotes the temperature. The positive pair $z_{1}$ and $z_{2}$, are two augmented views of the same image, while the negative samples, represented as $z_{k}$, encompass all other samples within the same training batch. 

So our loss function is the sum of the losses of two branches \ref{eq:eq3}. And we set $\lambda = 0.2$ to balance the scale of the two losses.

\begin{equation}
	\label{eq:eq2}
	\mathcal{L}_{c}=-\frac{1}{B}\sum_{i=1}^{B}\log\frac{e^{z_{2i}^T\cdot z_{2i+1}/t}}{\sum_{k=1}^{2B}e^{z_{2i}^T\cdot z_{k}/t}}
\end{equation}

\begin{equation}
    \label{eq:eq3}
    \mathcal{L}=\mathcal{L}_{reconstructive}+\lambda\cdot\mathcal{L}_{contrastive}
\end{equation}

\section{Experimental results and analysis}
To evaluate the quality of the learned representations, we performed two experiments on the three common used datasets in agricultural diseases classification task \cite{ip102,plantvillage,flooding}. We first conducted linear probing, where we added a MLP layer to the output representations and exclusively train the MLP classification layer while keeping other parameters frozen. Then we conduct the fine-tuning step, as we fine-tuned all parameters for downstream classification tasks. Additionally, by setting the parameter $\lambda$, we compared the enhancement in the feature extraction capability of the CRE model using contrastive learning.

\subsection{Pre-training implemtation}
In this paper, the GPID-22 was used to pre-train the model. We maintain a consistent input image resolution of 256x256. We also set the output length of VQGAN tokenizer to 256 tokens. Following MAE\cite{mae}, our default augmentations include robust random crop and resize and random flipping. The pre-training involves base-size vision Transformers \cite{ViT}, namely ViT-B. We employ the AdamW optimizer to train the model for 1600 epochs, using a batch size of 2048 for ViT-B. A cosine learning rate schedule with an 80-epoch warmup is applied. The base learning rate is set to $1.5\times 10^{-4}$ for ViT-B, further scaled by batchsize/256. The environment of our experiments is implemented with Python 3.8 on Ubuntu 21.04 system with a server of 8 $\times$ A800 GPU.

\subsection{Image Classification}
\subsubsection{Linear probing}
Linear probing is a basic evaluation protocol for self-supervised learning. To compare our method with other state-of-the-art self-supervied models, we pretrain MAE \cite{mae}, MoCo v3 \cite{moco_v3} and SimCLR\cite{simclr} respectively on GPID-22. In comparison to state-of-the-art MIM methods, we initially set $\lambda = 0$, indicating the exclusion of the contrastive loss. As illustrated in Table \ref{tab:linear_probing}, CRE ($\lambda = 0$) demonstrates superior performance, surpassing MAE by $3.1\%$ in ViT-B for GPID-22 linear probe top-1 accuracy. Furthermore, in contrast to other contrastive models, our approach outperforms MoCo by $2.4\%$ and SimCLR by $1.2\%$ in terms of accuracy. Notably, we observed a $2.3\%$ improvement in CRE accuracy when $\lambda$ was increased from $0$ to $0.2$, underscoring the substantial enhancement brought about by the addition of the contrastive loss to the model.

\vspace{-0.4cm}
\begin{table}[H]
    \centering
    \caption{Top-1 accuracy of linear probing on GPID-22. $\lambda = 0$ denotes that the model does not use the contrastive loss.}
    \label{tab:linear_probing}
    \scalebox{1}{
        \setlength{\tabcolsep}{6mm}{
    \begin{tabular}{l|ccc}
    \toprule
    Methods & Model & Acc($\%$) & F1($\%$)\\
    \hline
    \emph{MIM methods} \\
    MAE\cite{mae}  & ViT-B & 70.4 & 69.8\\
    CRE ($\lambda = 0$) & ViT-B & 71.2 & 70.5 \\
    \hline
    \emph{Contrastive methods} \\
    SimCLR v2\cite{simclr} & ResNet50 & 72.3 & 71.5\\
    MoCo v3\cite{moco_v3} & ViT-B & 71.1 & 70.3 \\
    CRE ($\lambda = 0.2$) & ViT-B & \textbf{73.5} & \textbf{72.8} \\
    \bottomrule
    \end{tabular}}}
\end{table}
\vspace{-1cm}

\subsubsection{Fine-tuning}
We conduct a comparative analysis of our method against state-of-the-art pest and disease classification models, as presented in Table \ref{tab:fin-tuning}. Across the IP102 datasets, our proposed CRE model demonstrates accuracy levels of $76.17\%$, accompanied by F1 score of $75.47\%$. Given the enhanced feature extraction capabilities inherent in Transformer-based models, it is not surprising that our approach exhibits superior performance compared to CNN-based models such as VGGNet \cite{VGG}, EfficientNet \cite{effnet}, and ResNet \cite{resnet} on the IP102 dataset.

\vspace{-0.3cm}
\begin{table}[H]
    \centering
    \caption{Comparisons with state-of-the-art methods on different datasets.}
    \label{tab:fin-tuning}
        \scalebox{0.9}{
    \begin{tabular}{{l}|*{6}{c}}
      \toprule
      \multirow{1}*{Method} & \multicolumn{2}{c}{IP102} & \multicolumn{2}{c}{Plant Village} & \multicolumn{2}{c}{CCD} \\
      \cmidrule(lr){2-3}\cmidrule(lr){4-5}\cmidrule(lr){6-7}
       & Acc($\%$) & F1($\%$) & Acc($\%$) & F1($\%$) & Acc($\%$) & F1($\%$) \\
      \hline
      \emph{Supervised methods} \\
      VGGNet \cite{VGG} & 43.65 & 40.21 & 99.57 & 96.42 & 80.10 & 79.49 \\
      EfficientNet \cite{effnet} & 60.46 & 59.21 & 99.93 & 97.30 & 82.41 & 81.30 \\
      ResNet101 \cite{resnet} & 54.67 & 53.79 & 99.34 & 99.23 & 83.06 & 82.27  \\
      MIL-Guided \cite{MIL-Guided}& 69.53 & 69.01 & - & - & - & - \\
      Attention-based MIL-Guided \cite{Attention-based_MIL-Guided}  & 68.31 & 68.02 & - & - & - & - \\
      Supervised-IN1K ViT \cite{Supervised-IN1K_ViT}  & 72.47 & 72.04 & 99.63 & 97.02 & - & -  \\
      \hline
      \emph{Self-supervised methods} \\
      MoCo v3 \cite{moco_v3}  & 73.14 & 72.28 & 99.78 & 97.13 & 83.02 & 82.49  \\
      MAE \cite{mae}  & 73.43 & 72.75 & 99.85 & 97.21 & 83.32 & 82.76  \\
      LSMAE \cite{LSMAE}  & 74.69 & 74.36 & 99.93 & 97.31 & - & - \\
      \hline
      CRE ($\lambda = 0$) & 75.46 & 74.69 & 99.94 & 97.38 & 84.26 & 83.43 \\
      CRE ($\lambda = 0.2$) & \textbf{76.17} & \textbf{75.47} & \textbf{99.95} & \textbf{97.56} & \textbf{85.47} & \textbf{84.71}  \\
      \bottomrule
    \end{tabular}}
\end{table} 

In contrast to the weakly supervised MIL-Guided model \cite{MIL-Guided}, our approach exhibits notable performance enhancements, achieving an increase of $6.64\%$ in accuracy and a $6.46\%$ improvement in F1 score on the IP102 dataset. In comparison to the supervised-model Supervised-IN1K ViT \cite{Supervised-IN1K_ViT}, which utilizes supervised learning on ImageNet-1K, our accuracy surpasses theirs by $3.7\%$. Furthermore, it is noteworthy that the performance of these supervised-learning methods lags behind that of other supervised-learning models such as MoCo v3 \cite{moco_v3}, MAE \cite{mae}, and LSMAE \cite{LSMAE}. This observation underscores the efficacy of self-supervised methods in facilitating superior transferability to downstream tasks.

Furthermore, we demonstrate superior performance of our methodologies compared to conventional self-supervised approaches, specifically the MIM and contrastive-learning models \cite{moco_v3,mae,LSMAE}, achieving accuracy improvements of $3.03\%$, $2.74\%$, and $1.48\%$ respectively. Prior self-supervised learning approaches in image processing typically employ raw images as inputs to transformers. In contrast, the CRE methodology adopted in this study utilizes quantized semantic tokens for both input and reconstruction target modalities. This divergence in approach attributed to the observed results, suggesting that our method enables the entire network to function at a semantic level, avoiding the loss of low-level details and consequently facilitating the extraction of more effective representations.

Additionally, we conducted experiments on two widely used datasets, Plant Village and Chinese Cucumber Leaf Dataset (CCD) \cite{flooding}, commonly employed in pest and disease classification tasks. Throughout these experiments, it was observed that our proposed method, along with other transformer-based approaches, achieved significantly high accuracy within 15 fine-tuning epochs. Notably, CRE exhibited the highest accuracy among all methods, surpassing MAE and MoCo by $0.10\%/0.20\%$ and $0.17\%/0.26\%$ respectively.

To demonstrate the impact of contrastive learning, we fine-tune both CRE ($\lambda = 0$) and CRE ($\lambda = 0.2$) on downstream tasks. As shown in Table \ref{tab:fin-tuning}, the model we only uses the reconstructive loss to pre-train still inferior to the results of using a contrastive loss together. The result shows an improvement of $0.71\%/0.01\%/1.21\%$ in accuracy and $0.78\%/0.18\%/1.28\%$ in F1 score on the IP102/Plant Village/CCD, respectively.

In summary, our proposed method exhibits superior classification performance compared to state-of-the-art approaches for two primary reasons: (1) The reconstructive learning process employs a tokenizer facilitating the operation on semantic tokens, thereby enhancing the extraction of more refined representations. (2) The constructive loss serves as a regularization mechanism, mitigating the risk of the encoder learning shortcut solutions. Consequently, the CRE model presented in this study emerges as an effective strategy for enhancing the performance of pest and disease classification.

\section{Conclusion}
In response to the considerable limitations in both data quality and diversity within plant pest and disease research, we have developed an extensive dataset named GPID-22. This dataset encompasses images depicting 183 distinct types of pests and diseases across 22 plant species, amounting to 205,371 images distributed across 199 different classes. Notably, GPID-22 stands out as one of the largest and most diverse datasets in the field. To maximize the utility of our extensive pre-training dataset, we propose an advanced network named CRE. The architecture of CRE integrates Contrastive Learning and Masked Image Modeling (MIM). Drawing inspiration from prior generative models, CRE incorporates semantic tokens learned by a vector-quantized Generative Adversarial Network (GAN) at both inputs and outputs, along with masking techniques. Additionally, we enhance feature representation by introducing a contrastive loss to the encoder output. Subsequently, we conduct validation experiments on IP102, Plant Village, and CCD datasets, achieving state-of-the-art results. These outcomes underscore the effectiveness of our proposed method.

\textbf{Acknowledgement.}
This work was supported in part by the National Key Research and Development Program of China (2022YFD2001501). All of the mentioned
support is gratefully acknowledged. In addition, thanks for all the help of
the teachers and students of the related universities.

\bibliographystyle{splncs04}
\bibliography{bibliography}

\end{document}